\documentclass[letterpaper, 10 pt, conference]{ieeeconf}  

\IEEEoverridecommandlockouts                              

\usepackage{amsmath, amssymb, mathtools}
\usepackage[table,xcdraw]{xcolor}
\usepackage{graphicx}
\usepackage{subfigure}
\usepackage{svg}
\usepackage{booktabs}
\usepackage{multirow}
\makeatletter
\let\NAT@parse\undefined
\makeatother
\usepackage[colorlinks=true, citecolor=green, linkcolor=red]{hyperref}  
\usepackage{cite}
\usepackage{algorithm}
\usepackage{algorithmic}
\title{\LARGE \bf
DroneMOT: Drone-based Multi-Object Tracking Considering Detection Difficulties and Simultaneous Moving of Drones and Objects}

\author{Peng Wang, Yongcai Wang and Deying Li 
\thanks{All authors are with the Department of Computer Science, School of Information, Renmin University of China, Beijing 100872, China. Corresponding author: Yongcai Wang. Email \{{\tt\small peng.wang, ycw, deyingli\}@ruc.edu.cn}}%
\thanks{Dr. Li is supported in part by the National Natural Science Foundation of China Grant No. 12071478. 
Dr. Wang is supported in part by the National Natural Science Foundation of China Grant No. 61972404, Public Computing Cloud, Renmin University of China, and the Blockchain Lab, School of Information, Renmin University of China.}
}

\begin{document}

%

\maketitle
\thispagestyle{empty}
\pagestyle{empty}

\begin{abstract}
Multi-object tracking (MOT) on static platforms, such as by surveillance cameras, has achieved significant progress, with various paradigms providing attractive performances. However, the effectiveness of traditional MOT methods is significantly reduced when it comes to dynamic platforms like drones. This decrease is attributed to the distinctive challenges in the MOT-on-drone scenario: (1) objects are generally small in the image plane, blurred, and frequently occluded, making them challenging to detect and recognize; (2) drones move and see objects from different angles, causing the unreliability of the predicted positions and feature embeddings of the objects. 
This paper proposes DroneMOT, which firstly proposes a Dual-domain Integrated Attention (DIA) module that considers the fast movements of drones to enhance the drone-based object detection and feature embedding for small-sized, blurred, and occluded objects.
Then, an innovative Motion-Driven Association (MDA) scheme is introduced, considering the concurrent movements of both the drone and the objects. Within MDA, an Adaptive Feature Synchronization (AFS) technique is presented to update the object features seen from different angles. Additionally, a Dual Motion-based Prediction (DMP) method is employed to forecast the object positions. Finally, both the refined feature embeddings and the predicted positions are integrated to enhance the object association.
Comprehensive evaluations on VisDrone2019-MOT and UAVDT datasets show that DroneMOT provides substantial performance improvements over the state-of-the-art in the domain of MOT on drones. 
\end{abstract}

\section{INTRODUCTION}
Multi-object tracking (MOT) is a critical task in computer vision, which has a wide range of applications in autonomous driving\cite{UniAD} and video surveillance\cite{oh2011large}. 
The goal of MOT is to find the trajectories of objects through continuous observations by cameras. 
MOT methods can be broadly categorized into two paradigms: tracking-by-detection\cite{TbD, TbDsurvey,centertrack, JDE, sort} and tracking-by-regression\cite{wan2021tracking,cai2022memot,meinhardt2022trackformer}. 
Currently, due to the great success of deep learning-based object detection\cite{he2016deep,fasterRcnn,centernet}, tracking-by-detection methods\cite{SMILEtrack}\cite{TbDsurvey}, which firstly detect objects in each frame and then associate the detections with the trajectories, have a leading performance in MOT.

\begin{figure}
        \centering
        \setlength{\abovecaptionskip}{-0.4cm}
        \includegraphics[width=3.4in]{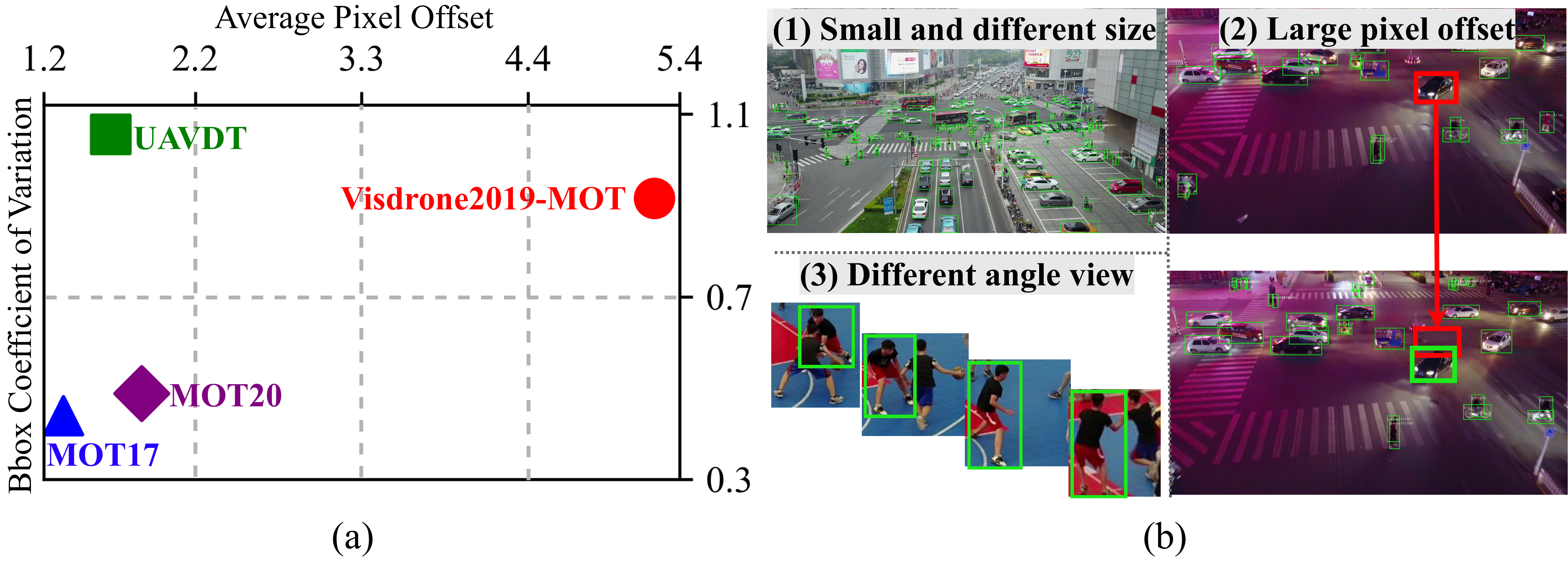} 
        \caption{\textbf{Challenges of MOT on drones.} 
        (a) comparisons between conventional MOT datasets(MOT17/20) and drone-based MOT datasets(Visdrone2019-MOT and UAVDT). 
        The x-axis represents the average change in the pixel position of the same object in adjacent frames. In contrast, the y-axis represents the coefficient of variation (variance/mean) of the object's bbox size. 
        (b) Visualization of these challenges, encompassing small-scale objects, large pixel offsets, and varying angle views.} 
        \label{Challenges}
        \vspace{-0.6cm}
\end{figure}

MOT has shown impressive performance for static cameras\cite{mot15,mot16,mot20}. 
However, when applied to drones or unmanned aerial vehicles, the performance of existing MOT methods decreases significantly\cite{zhu2021detection}. This decrease in performance is attributed to the difficulty in accurately detecting objects and associating them with their trajectories. These challenges are inherent to MOT-on-drone scenarios, as illustrated in Fig. \ref{Challenges}. 
At first, the elevated altitude at which drones operate often results in smaller apparent scales of the objects in the footage. Additionally, the swift movement of drones can introduce motion blur and occlusion into the video frames. 
Combining these factors makes it challenging to detect objects and extract meaningful feature embeddings\cite{mueller2016benchmark}\cite{kalra2019dronesurf}.
Furthermore, when the drone and the objects move simultaneously, there can be significant shifts in the pixel positions of the same object across consecutive frames. Such irregular movement might also cause objects near the camera's edge to appear discontinuously. With the drone in motion, the same object can be viewed from multiple angles, leading to inconsistent features.
Therefore, data association based on the coherence of target pixel positions and the consistency of target features tends to perform poorly under the dynamic conditions of drones.

Given the significant importance of drone-based object detection and tracking in various applications\cite{mueller2016benchmark}\cite{abdulrahim2016traffic}, several methods have emerged. 
One dominant approach adheres to the tracking-by-detection paradigm, emphasizing enhancements in drone-based object detection and feature embedding.  
For instance, Wang et al.\cite{wang2019orientation} modified YOLOv3\cite{redmon2018yolov3} to utilize three different resolution feature maps for vehicle detection and tracking in UAV videos.
UAVMOT\cite{uavmot} leverages the correlation layer between two adjacent frames to reinforce ID embedding based on features.
Some methods have been reported to address association issues.
Zhang et al.\cite{zhang2019eye} employ the TNT network\cite{wang2019exploit} for detection and directly calculate the Semi-Direct Visual Odometry by Multi-View Stereo for data association.
Other studies\cite{schreiber2019advanced,hosseinpoor2016pricise,wang2019development} utilize RTK, IMU, or GPS to directly compute the drone's poses, aiming to boost the performance of drone MOT. However, these methods require additional equipment.

In this work, we rely solely on the image information and propose \textbf{DroneMOT}, which not only enhances object detection and feature embedding but also considers simultaneous motions of the drone and the objects to improve the \textbf{robustness} of the data-association.
In particular, in the detection module, we introduce a \textit{Dual-Domain Integrated Attention (DIA)}, which integrates Spatial Attention and Heatmap-Guided Temporal Attention to achieve more accurate and comprehensive detections with embedding.
In the data-association module, we propose an innovative \textit{Motion-Driven Association (MDA)} scheme considering the simultaneous movement of the drone and the objects.
In MDA, we first present a \textit{Adaptive Feature Synchronization (AFS)} module that refines trajectory appearance by dynamically adjusting the feature weights based on the detection scores and preserving key historical features from different angles of the same object. 
Then, we introduce the \textit{Dual Motion-based Prediction (DMP)} module. Instead of solely focusing on the target motion, DMP also takes the drone motion into account. We decompose the drone's motion into three primary components: hovering, translation, and rotation. Combining the motion of the drone and the motions of the objects, the trajectory's pixel position in the subsequent frame is more accurately predicted.
The key contributions are summarized as follows:

\begin{itemize}
\item Dual-Domain Integrated Attention (DIA) is proposed to enhance the detection and feature embedding of small-sized, blurred, and occluded objects in videos captured by drone. 
\item Motion-Driven Association (MDA) is proposed for robust data association, which includes AFS to refine the trajectory appearance and DMP to predict the object position considering the simultaneous motions of the drone and the objects.
\item Extensive evaluations on the Visdrone2019-MOT\cite{visdrone} and UAVDT\cite{UAVDT} datasets demonstrate that DroneMOT outperforms the state-of-the-art methods for multi-object tracking on drones.
\end{itemize}

\begin{figure*}[htbp]
        \centering  
        \setlength{\abovecaptionskip}{-0.4cm}
        \includegraphics[width=7in]{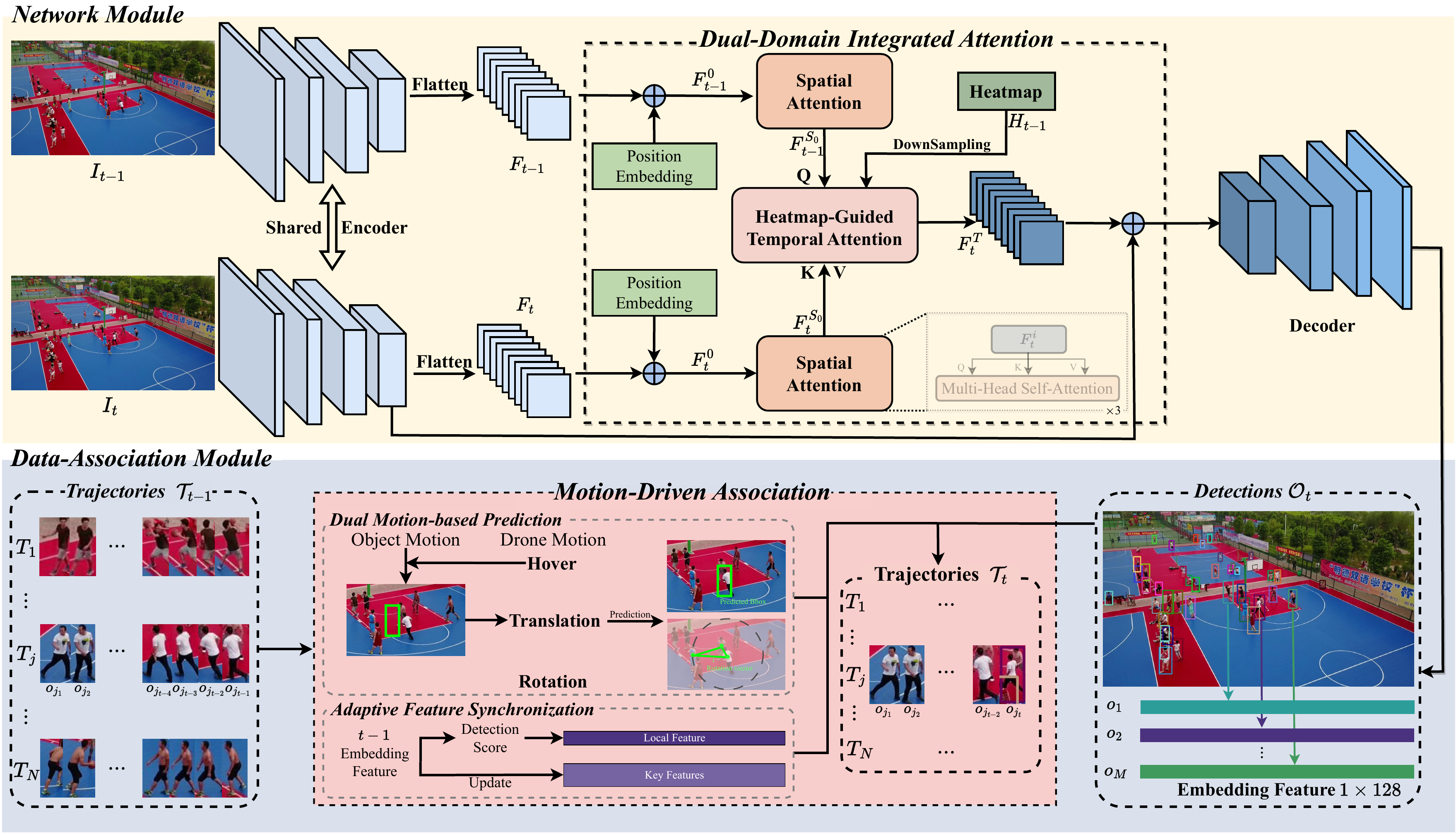}  
        \caption{\textbf{The overall architecture of DroneMOT.} 
        It primarily consists of two modules: the network module (\ref{NM}) for online detection and feature embedding and the data-association module (\ref{MDA}) to associate detections with stored trajectories of objects.}  
        \label{fig:network}  
        \vspace{-0.6cm}
\end{figure*}

\section{Related Work}
\noindent
\textbf{Multi-Object-Tracking on Drone. } 
MOT algorithms are usually divided into tracking-by-detection paradigms\cite{sort,deepsort, JDE, TrackRCNN,fairmot,uniconrn} and tracking-by-regression paradigms\cite{centertrack, TbR_bells,qin2023motiontrack,zeng2022motr,chu2023transmot,cai2022memot,liu2023uncertainty}. 
Due to the unpredictable and irregular properties of the simultaneous movement of drones and objects, MOT for drones \cite{huang2021multiple,zhang2019eye} typically adopts the tracking-by-detection paradigm. This approach first uses a network to detect objects in each frame and then associates these detections with the stored trajectories.
PAS Tracker\cite{stadler2020pas} uses an additional ReID network to obtain object features and combines position, appearance, and size information jointly to make full use of the object representations. 
UAVMOT\cite{uavmot} utilizes the correlation layer\cite{li2019siamrpn++,li2018high,bhat2019learning,cao2022tctrack} between two adjacent frames to strengthen the embedding features, and develops an adaptive motion filter to complete the object ID association accurately. 
GLOA\cite{shi2023global} proposes a global-local awareness detector to extract scale variance feature information from the input frames for the frequent occluded objects.
FOLT\cite{yao2023folt} adopts a light-weight optical flow extractor to extract object detection features and motion features at a minimum cost to improve the detection of small objects.
Although some research has begun to focus on data association, the drone-based MOT methods are still focused on building powerful detectors. 
In this work, we present an integrated framework tailored not only for the enhanced detection of small, blurred, and occluded objects but also for data-association strategies specifically designed to accommodate the motion of drones.

\noindent
\textbf{Data-Association. }
Early MOT approaches, such as SORT\cite{sort,deepsort}, adopt the data-association method. These methods employ a Kalman filter\cite{kalman1960new,julier1997new,gustafsson2002particle,smith1962application} to predict an object's trajectory position in the subsequent frame, serving as the motion model. Concurrently, a network\cite{deepsort_network} is utilized to obtain the object feature embedding, acting as the appearance model. By integrating both the motion and the appearance models, data-association is achieved using the Hungarian algorithm\cite{kuhn1955hungarian}.
BoT-SORT\cite{botsort} utilizes an enhanced Kalman filter and compensates for camera motion to achieve a more accurate motion model. 
OC-SORT\cite{oc-sort} uses object observations to compute a virtual trajectory to correct the error accumulation of the Kalman filter during the occlusion period.
Meanwhile, some researchers have focused on the appearance model to get effective and comprehensive features. 
CorrTracker\cite{CorrTracker} uses the correlation layer\cite{dosovitskiy2015flownet} to calculate the spatio-temporal correlation of features between adjacent frames, thereby obtaining more accurate object feature embedding.
GHOST\cite{GHOST} analyzes MOT failure cases and proposes a combination method of proxy appearance features with a simple motion model, leading to strong tracking results.
In addition, ByteTrack\cite{bytetrack} employs a multi-level data-association method. The trajectories are first matched with the detections that have high detection scores, and the remaining trajectories are matched with the detections that have low detection scores.
In this work, we adopt these advanced data-association methods and further consider the motion patterns of drones to specifically design motion and appearance models for data association on drones.

\section{METHOD}
\noindent
\textbf{DroneMOT} is primarily split into two modules: the network module (\ref{NM}) for detection and feature embedding, and the data-association module (\ref{MDA}) based on the result of the network module. The image $I_t \in \mathbb{R}^{W \times H \times 3} $ captured by the moving drone at the $t$-th frame is fed into the network along with the previous frame image $I_{t-1}$. The results of the network module, represented by $\mathcal{O}_{t} = \left\{o_1, o_2, \cdots, o_i, \cdots, o_M \right\} $ consist of $M$ detections where $o_i = \left(b_i, s_i, f_i\right)$. 
Here,  $b_i$ represents the bounding box $\left(x, y, w, h\right)$, $s_i$ is the detection score, and $f_i$ is the feature embedding vectors. The data association module takes the detections $\mathcal{O}_{t}$ and all $N$ stored trajectories of the objects $\mathcal{T}_{t-1} = \left\{T_1, T_2, \cdots, T_j, \cdots, T_N \right\}$ as inputs, where $T_j = \left\{o_{j_1}, o_{j_3}, \cdots, o_{j_{t-1}} \right\}$, and $o_{j_{t-1}}$ represents the detection associated with the trajectory $j$ in the $t-1$-th frame. The goal of the data association module is to match each detection with a trajectory, treat the unmatched detections as the new trajectories, and ultimately produce the final tracking results $\mathcal{T}_{t}$. An overview of the proposed \textbf{DroneMOT} is presented in Fig. \ref{fig:network}.

\subsection{Network Module} \label{NM}
In the network module, we utilize the DLA34\cite{dla34} network as the backbone, which is an encoder-decoder architecture. The encoder uses shared convolution layers to extract local features from images $\left\{I_{t-1}, I_t\right\}$. After flattening the local features, these features denoted as $\left\{F_{t-1}, F_t\right\}$ serve as inputs to the \textbf{Dual-Domain Integrated Attention (DIA)} module. DIA module consists of two parts: \textit{Spatial Attention} and \textit{Heatmap-Guided Temporal Attention}.

\noindent
\textbf{Spatial Attention.} 
The Spatial Attention layer aims to augment object features with spatial positional information and the relationships between objects, enabling the network to distinguish different small-scale objects easily. 
The effectiveness of the Spatial Attention is illustrated in Fig. \ref{fig:DIA_visual}(a). 
To achieve this goal, we firstly add the flattened local features $F_{t-1}, F_t$ with a 2D extension of the standard position encoding\cite{detr} to make the features cognizant of their global positions within the entire 2D image feature space: 
\begin{equation}
\begin{aligned}
F^0_t &= F_t + \text{PosEncod}.
\end{aligned}
\end{equation}
Then we adopt three multi-head self-attention layers separately to enhance the spatial relationships and object interactions within the feature maps, thereby crafting a more spatially aware representation feature $F^{S_0}_{t-1}, F^{S_0}_{t}$:
\begin{equation}
\begin{aligned}
F^{i+1}_t = \text{Norm}(F^i_t &+ \text{MultiHead}(F^i_t,F^i_t,F^i_t)),i = 0, 1, \\
F^{S_0}_t = \text{Norm}&(F^2_t + \text{MultiHead}(F^2_t,F^2_t,F^2_t)).
\end{aligned}
\end{equation}
where $t$ can be replaced by $t-1$, ``MultiHead" refers to the multi-head attention\cite{transformer} following the query, key, and value, and ``Norm" represents the layer normalization.

\noindent
\textbf{Heatmap-Guided Temporal Attention.}
The temporal attention layer focuses on the evolution of features for the same object over successive time steps. 
In aerial tracking, the presence of motion blur or occlusion often leads to ineffective temporal contexts.
To filter out the regions without objects and to heighten the feature's focus on the objects affected by motion blur and occlusion, we propose to use the heatmap of the $t-1$-th frame as the filter's attention. 
As illustrated in Fig. \ref{fig:DIA_visual}(b)(c), this heatmap-guided filter leads to a more context-aware interpretation of the blurred and occluded objects detected from the visual sequence. 

\begin{figure}[t] 
        \centering
        \setlength{\abovecaptionskip}{-0.6cm}
        \includegraphics[width=3.4in]{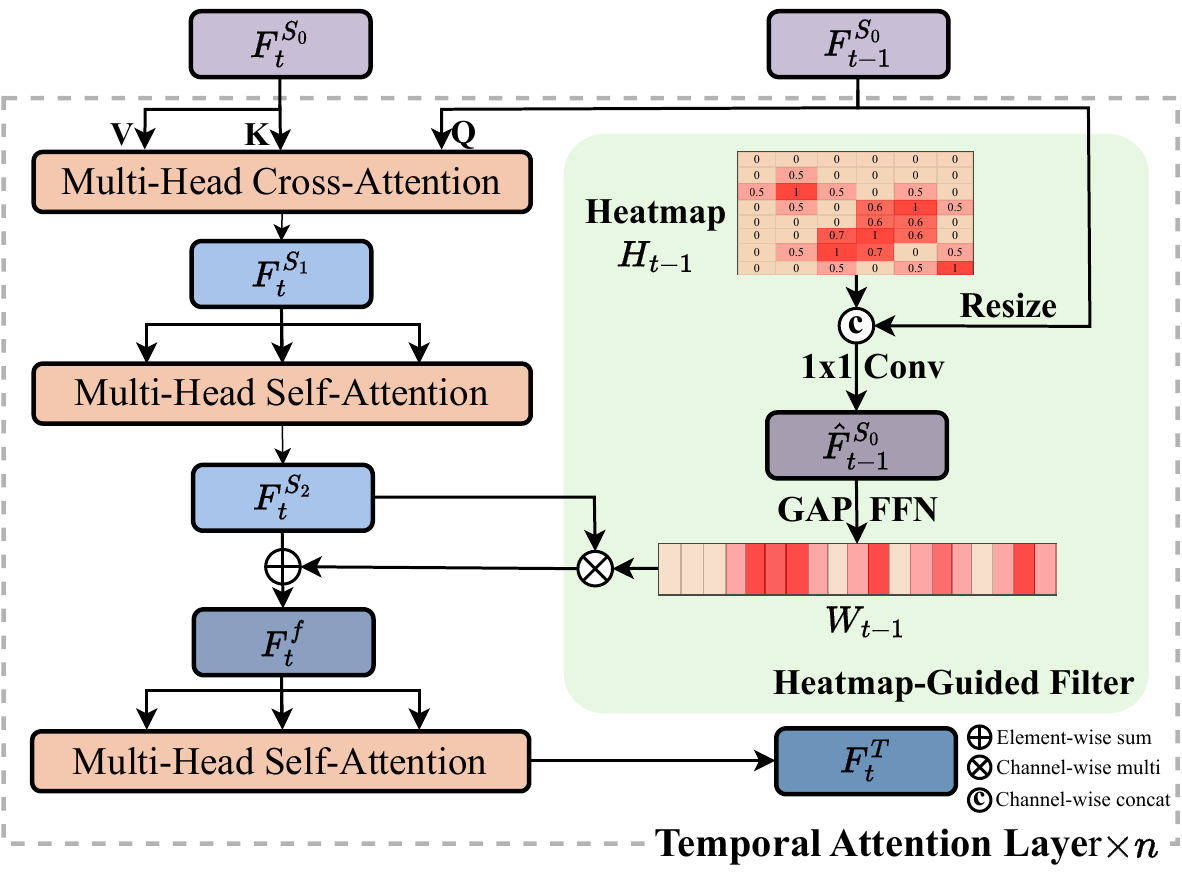}
        \caption{Structure of Heatmap-Guided Temporal Attention.} 
        \label{fig:DIA}
        \vspace{-0.6cm}
\end{figure}

Specifically, given the adjacent spatial enhanced feature $F^{S_0}_{t-1}, F^{S_0}_t$, and the heatmap $H_{t-1}$ obtained from the $t-1$-th frame, we acquire the output feature $F^{S_2}_t$ of the stacked multi-head attention layer in the $t$-th frame:
\begin{equation}
\begin{aligned}
F^{S_1}_t &= \text{Norm}(F^{S_0}_t + \text{MultiHead}(F^{S_0}_{t-1},F^{S_0}_t,F^{S_0}_t)), \\
F^{S_2}_t &= \text{Norm}(F^{S_1}_t + \text{MultiHead}(F^{S_1}_{t},F^{S_1}_t,F^{S_1}_t)).
\end{aligned}
\end{equation}
Then as presented in Fig. \ref{fig:DIA}, the feature representation $\hat{F}^{S_0}_{t-1}$ is generated by concatenating the heatmap with the resized convolutional features, followed by a 1$\times$1 convolution.
A heatmap-guided weight $W_{t-1}$ is derived via Global Average Pooling (GAP) and a feed-forward network (FFN). This weight is then multiplied with the feature $F^{S_2}_t$, creating a refined feature representation $F^f_t$ guided by the heatmap. Finally, $F^T_t$ is obtained by the multi-head attention:
\begin{equation}
\begin{aligned}
\hat{F}^{S_0}_{t-1} = \mathcal{F}(\text{Cat}&(H_{t-1},\text{Resize}(F^{S_0}_{t-1}))), \\
W_{t-1} = \text{FFN}&(\text{GAP}(\hat{F}^{S_0}_{t-1})), \\
F^f_t = F^{S_2}_t + &F^{S_2}_t \times W_{t-1}, \\
F^{T}_t = \text{Norm}(F^{f}_t + &\text{MultiHead}(F^{f}_{t},F^{f}_t,F^{f}_t)).
\end{aligned}
\end{equation}
where $\mathcal{F}$ represents a convolution layer, and $\text{FFN}$ means a feed-forward network.

The local feature $F_t$ at the $t$-th frame, combined with the results $F^T_t$ from the DIA module, is utilized as the input to the decoder, resulting in the Detection Head. 
Following \cite{fairmot}, the detection head applies successive convolutional operations to obtain the heatmap $H_t$ of the objects, which can be used as the input to the network of the $t+1$-th frame, along with the corresponding width, height, and feature embedding. 
These form the object detection results and their feature embeddings, i.e., $\mathcal{O}_{t} = \left\{o_1, o_2, \cdots, o_M \right\}$ for the $t$-th frame.

\subsection{Motion-Driven Association} \label{MDA}
\textbf{Motion-Driven Association} (MDA) takes detections $\mathcal{O}_{t}$ in the $t$-th frame  and trajectories $\mathcal{T}_{t-1}$ from the $t-1$-th frame  as inputs. 
Considering the simultaneous movements of both the drone and the objects, MDA consists of two primary components: (1) Adaptive Feature Synchronization (AFS) and (2) Dual Motion-based Prediction(DMP). Finally, both the refined feature embeddings and the precise predicted positions are integrated to enhance the object association to get the trajectory $\mathcal{T}_{t}$ for the $t$-th frame.

\begin{table*}[htbp]
\centering
\caption{Quantitative comparisons between DroneMOT and other methods on VisDrone2019-MOT test-dev and UAVDT test set. Methods in \textcolor[RGB]{116, 200, 235}{blue} block are MOT methods specifically for the drone. The best results are marked in \textbf{bold}.}
\label{table:Quantitative comparisons}
\resizebox{0.98\textwidth}{!}{%
\begin{tabular}{lll|llllllll}
\hline
\rowcolor[HTML]{C0C0C0} 
Dataset                            & Method                                           & Pub\&Year                           & IDF1$\uparrow$                        & MOTA$\uparrow$                        & MOTP$\uparrow$                        & MT$\uparrow$                         & ML$\downarrow$                       & FP$\downarrow$                         & FN$\downarrow$                         & IDs$\downarrow$                      \\ \hline
                                   & SiamMOT\cite{shuai2021siammot}             & CVPR2021                            & 48.3                                  & 31.9                                  & 73.5                                  & -                                  & -                                  & 24123                                  & 142303                                 & 862                                 \\
                                   & MOTR\cite{zeng2022motr}                          & ECCV2022                            & 41.4                                  & 22.8                                  & 72.8                                  & 272                                  & 825                                  & 28407                                  & 147937                                 & 959                                  \\
                                   & ByteTrack\cite{bytetrack}                        & ECCV2022                            & 40.8                                  & 25.1                                  & 72.4                                  & 446                                  & 1099                                 & 34044                                  & 194984                                 & 1590                                 \\
                                   & OC-SORT\cite{oc-sort}                            & CVPR2023                            & 50.4                                  & 39.6                                  & 73.3                                  & -                                     &-                                      & \textbf{14631}                                  & 123513                                 & 986                                  \\
                                   & STDFormer\cite{hu2023stdformer}                  & TCSVT2023                           & 57.1                                  & \textbf{45.9}                                  & 77.9                                  & 684                                  & 538                                  & 21288                                  & 101506                                 & 1440                                 \\ \cline{2-11} 
                                   & \cellcolor[HTML]{E3F4FB}UAVMOT\cite{uavmot}      & \cellcolor[HTML]{E3F4FB}CVPR2022    & \cellcolor[HTML]{E3F4FB}51            & \cellcolor[HTML]{E3F4FB}36.1          & \cellcolor[HTML]{E3F4FB}74.2          & \cellcolor[HTML]{E3F4FB}520          & \cellcolor[HTML]{E3F4FB}574          & \cellcolor[HTML]{E3F4FB}27983          & \cellcolor[HTML]{E3F4FB}115925         & \cellcolor[HTML]{E3F4FB}2775         \\
                                   & \cellcolor[HTML]{E3F4FB}FOLT\cite{yao2023folt}   & \cellcolor[HTML]{E3F4FB}MM2023      & \cellcolor[HTML]{E3F4FB}56.9          & \cellcolor[HTML]{E3F4FB}42.1          & \cellcolor[HTML]{E3F4FB}\textbf{77.6}          & \cellcolor[HTML]{E3F4FB}-             & \cellcolor[HTML]{E3F4FB}-             & \cellcolor[HTML]{E3F4FB}24105          & \cellcolor[HTML]{E3F4FB}107630         & \cellcolor[HTML]{E3F4FB}\textbf{800}          \\
                                   & \cellcolor[HTML]{E3F4FB}GLOA\cite{shi2023global} & \cellcolor[HTML]{E3F4FB}J-STARS2023 & \cellcolor[HTML]{E3F4FB}46.2          & \cellcolor[HTML]{E3F4FB}39.1          & \cellcolor[HTML]{E3F4FB}76.1          & \cellcolor[HTML]{E3F4FB}581          & \cellcolor[HTML]{E3F4FB}824          & \cellcolor[HTML]{E3F4FB}18715          & \cellcolor[HTML]{E3F4FB}158043         & \cellcolor[HTML]{E3F4FB}4426         \\
\multirow{-9}{*}{VisDrone2019-MOT} & \cellcolor[HTML]{E3F4FB}DroneMOT                 & \cellcolor[HTML]{E3F4FB}Ours        & \cellcolor[HTML]{E3F4FB}\textbf{58.6}          & \cellcolor[HTML]{E3F4FB}43.7          & \cellcolor[HTML]{E3F4FB}71.4          & \cellcolor[HTML]{E3F4FB}\textbf{689}          & \cellcolor[HTML]{E3F4FB}\textbf{397}          & \cellcolor[HTML]{E3F4FB}41998          & \cellcolor[HTML]{E3F4FB}\textbf{86177}          & \cellcolor[HTML]{E3F4FB}1112         \\ \hline
                                   & DeepSORT\cite{deepsort}                          & ICIP2017                            & 58.2                                  & 40.7                                  & 73.2                                  & 595                                  & 338                                  & 44868                                  & 155290                                 & 2061                                 \\
                                   & SiamMOT\cite{shuai2021siammot}             & CVPR2021                            & 61.4                                  & 39.4                                  & 76.2                                  & -                                  & -                                  & 46903                                  & 176164                                 & 190                                 \\
                                   & ByteTrack\cite{bytetrack}                        & ECCV2022                            & 59.1                                  & 41.6                                  & 79.2                                  & -                                  & -                                 & \textbf{28819}                                  & 189197                                 & 296                                 \\
                                   & OC-SORT\cite{oc-sort}                            & CVPR2023                            & 64.9                                  & 47.5                                  & 74.8                                  &-                                      &-                                      & 47681                                  & 148378                                 & 288                                  \\ \cline{2-11} 
                                   & \cellcolor[HTML]{E3F4FB}UAVMOT\cite{uavmot}      & \cellcolor[HTML]{E3F4FB}CVPR2022    & \cellcolor[HTML]{E3F4FB}67.3          & \cellcolor[HTML]{E3F4FB}46.4          & \cellcolor[HTML]{E3F4FB}72.7          & \cellcolor[HTML]{E3F4FB}624          & \cellcolor[HTML]{E3F4FB}221          & \cellcolor[HTML]{E3F4FB}66352          & \cellcolor[HTML]{E3F4FB}115940         & \cellcolor[HTML]{E3F4FB}456          \\
                                   & \cellcolor[HTML]{E3F4FB}FOLT\cite{yao2023folt}   & \cellcolor[HTML]{E3F4FB}MM2023      & \cellcolor[HTML]{E3F4FB}68.3          & \cellcolor[HTML]{E3F4FB}48.5          & \cellcolor[HTML]{E3F4FB}\textbf{80.1} & \cellcolor[HTML]{E3F4FB}-             & \cellcolor[HTML]{E3F4FB}-             & \cellcolor[HTML]{E3F4FB}36429 & \cellcolor[HTML]{E3F4FB}155696         & \cellcolor[HTML]{E3F4FB}338          \\
                                   & \cellcolor[HTML]{E3F4FB}GLOA\cite{shi2023global} & \cellcolor[HTML]{E3F4FB}J-STARS2023 & \cellcolor[HTML]{E3F4FB}68.9          & \cellcolor[HTML]{E3F4FB}49.6          & \cellcolor[HTML]{E3F4FB}79.8          & \cellcolor[HTML]{E3F4FB}626          & \cellcolor[HTML]{E3F4FB}220          & \cellcolor[HTML]{E3F4FB}55822          & \cellcolor[HTML]{E3F4FB}115567         & \cellcolor[HTML]{E3F4FB}433          \\
\multirow{-8}{*}{UATDT}            & \cellcolor[HTML]{E3F4FB}DroneMOT                 & \cellcolor[HTML]{E3F4FB}Ours        & \cellcolor[HTML]{E3F4FB}\textbf{69.6} & \cellcolor[HTML]{E3F4FB}\textbf{50.1} & \cellcolor[HTML]{E3F4FB}74.5          & \cellcolor[HTML]{E3F4FB}\textbf{638} & \cellcolor[HTML]{E3F4FB}\textbf{178} & \cellcolor[HTML]{E3F4FB}57411          & \cellcolor[HTML]{E3F4FB}\textbf{112548} & \cellcolor[HTML]{E3F4FB}\textbf{129} \\ \hline
\end{tabular}%
}
\vspace{-0.6cm}
\end{table*}

\noindent
\textbf{Adaptive Feature Synchronization.}
In previous work \cite{fairmot,deepsort}, the appearance feature vectors of a trajectory only consider the local feature, which is updated by an Exponential Moving Average (EMA) of the current feature vector and the historical feature vector. EMA typically requires a fixed weight coefficient $\alpha$ to control the contribution of the historical feature vectors.

As an appearance model for data-association, AFS categorizes the features of trajectories into local and key features. 
To obtain more accurate local features, we dynamically adjust the weight coefficient $\alpha$ based on the detection score of the current frame. 
In addition, to address scenarios with sudden changes in target angles or extended occlusions, we preserve a subset of historical features as key features.

For the local feature, we use the detection score $s_t$ as the proxy to dynamically adjust the weight coefficient $\alpha$ in EMA, which is defined as
\begin{equation}
\begin{aligned}
f^{local}_t = \alpha f^{local}_{t-1} &+ (1-\alpha) f_t,  \\
\alpha = \alpha_{f} + (1&-\alpha_{f})e^{(\theta-s_t)} . 
\end{aligned}
\end{equation}
where $\alpha_{f}$ is a fixed value, usually set to 0.9, $s_t$ represents the object detection score, and $\theta$ is a detection confidence threshold to filter out noisy detections. 
For high-confidence detections, $\alpha$ approaches $\alpha_{f}$, increasing its impact on the local feature.

As for the key features, AFS retains a portion of historical features for every trajectory. The key features are typically updated by employing the least recently used algorithm to store the ten key features. 

\noindent
\textbf{Dual Motion-based Prediction.}
Unlike existing methods such as \cite{fairmot, botsort} that only consider the movement of objects, DMP also incorporates the drone's motion. 
We classify the drone's movements into three fundamental types: hovering, translation, and rotation. 
When the drone hovers, the camera can be approximated as a fixed camera. We can utilize the Kalman filter\cite{kalman1960new} for fixed cameras to predict the trajectories $\mathcal{T}_{t-1}$'s position in the $t$-th frame. 
When the drone undergoes translation or rotation, we compensate separately for the movements of the drone to improve the object-trajectory association.

Regarding the translation, following \cite{botsort}, we calculate the affine matrix between two frames and subsequently determine the position of the trajectories after the affine transformation. This method, termed Camera Motion Compensation, effectively compensates for the impact of translation of the drone on MOT. 
For rotation, we observed that the shape of the triangle formed by the object and its surrounding objects in adjacent frames is almost congruent. 
Therefore, the rotation vector of an object can be effectively captured using the intrinsic features of a triangle: 
$v_t = [\alpha_t, \beta_t, l_t]$ for an object in the $t$-th frame. Here, $\alpha, \beta$ denote the two smallest angles of the triangle, while $l$ represents the side length opposite the largest angle. The triangle is formed by the object, the farthest object, and the nearest object within a radius of $R$ pixels.
 
Finally, by integrating the drone's hovering and translation with the objects' movement, we can predict the trajectories' positions in the $t$-th frame. This integration enables us to compute the IOU cost matrix $I_C$ between the predicted object positions (bounding box with positions) and the detected object positions. 
Moreover, we evaluate the cosine similarity between the rotation vector of the trajectories and that of the detections, resulting in the rotation cost matrix $R_C$. 
On the other hand, the AFS module efficiently calculates the appearance cost matrix $A_C$ based on the minimal cosine value discerned between the feature of the detections and both the local feature and the key features of the trajectories. 
Therefore, the final cost matrix is typically formulated by combining the three cost matrices, represented as: 
\begin{equation}
C = I_C + w_aA_C + w_rR_C
\label{eqn:loss}
\end{equation}
By using a linear sum assignment\cite{kuhn1955hungarian}, each detection can uniquely correspond to a trajectory. Unmatched targets are treated as new trajectories, yielding the trajectories $\mathcal{T}_{t}$ for the $t$-th frame.

\section{EXPERIMENTS}
\subsection{Experimental Setup}
\noindent
\textbf{Dataset.} 
We evaluate the proposed methods using two multi-object tracking datasets for drones: (1) VisDrone2019-MOT\cite{visdrone} and (2) UAVDT\cite{UAVDT}. They are both developed for multi-category tracking using drones. 
The VisDrone2019-MOT dataset \cite{visdrone} is divided into three parts: a training set (56 sequences), a validation set (7 sequences), and a test set (33 sequences). It encompasses ten categories: pedestrian, person, car, van, bus, truck, motor, bicycle, awning-tricycle, and tricycle.
The UAVDT dataset \cite{UAVDT} is explicitly designed for vehicle object tracking. It is split into two parts: a training set and a test set, covering three categories: car, truck, and bus. The video images in this dataset offer a resolution of 1080 × 540 pixels and showcase various illumination conditions, including sunshine, fog, and rain.

\noindent
\textbf{Metrics.} 
We adopt IDF1\cite{ristani2016performance}, MOTA\cite{bernardin2008evaluating}, and ID switching (IDs)\cite{bernardin2008evaluating} as the primary evaluation metrics to evaluate our proposed DroneMOT with other state-of-the-arts approaches. 
MOTA is computed based on FP, FN, and IDs, which focus more on the detection performance. And IDF1 evaluates the identity association accuracy of the tracking results.

\noindent
\textbf{Training Details.} 
We train DroneMOT for 30 epochs on six NVIDIA GeForce RTX 2080ti GPUs with batch size 12. 
In the multiple loss functions, we modify the EQ-Loss v2\cite{tan2021equalization} to supervise the heatmap. Furthermore, L1 loss and Triplet loss\cite{dong2018triplet} are separately used to deal with the width and height of the object and the object ID.

\noindent
\textbf{Tracking Details.}
At the data-association stage, we follow ByteTrack\cite{bytetrack} to set the high detection score threshold to 0.6 and the low detection score threshold to 0.1.
In Dual Motion-based Prediction, $w_a, w_r$ in Equation.~\ref{eqn:loss} are set to 0.5 and 0.1, respectively. Furthermore, $R$ in AFS module is set to 100 pixels. 

\subsection{Comparison with the state-of-the-art methods}
We compare DroneMOT with state-of-the-art (SOTA) trackers, including those specifically tailored for MOT on drones including UAVMOT\cite{uavmot}, FOLT\cite{yao2023folt}, GLOA\cite{shi2023global}
and the generic ones including SiamMOT\cite{shuai2021siammot}, MOTR\cite{zeng2022motr}, ByteTrack\cite{bytetrack}, OC-SORT\cite{oc-sort}, and STDFormer\cite{hu2023stdformer}. 
The performance results on the two drone-based MOT datasets are presented in the following sections.

\noindent
\textbf{Visdrone2019-MOT.}
In this dataset, we train using all categories. However, we adhere to the official VisDrone toolkit for evaluation, which focuses on five categories: car, bus, truck, pedestrian, and van—consistent with other trackers. Our results on the VisDrone2019 test-dev set are presented in Table \ref{table:Quantitative comparisons}. DroneMOT stands out, achieving the highest IDF1 score of 58.6\%, which is a marked improvement over competing methods. This score underscores DroneMOT's effectiveness in correctly identifying and matching object identities. 
Furthermore, DroneMOT excels in detection capabilities, recording the lowest FN count of 86,177. 
Moreover, it boasts the highest MT while registering the fewest ML, emphasizing its precision and consistency in maintaining trajectory IDs.

\noindent
\textbf{UAVDT.}
The UAVDT dataset presents a more pronounced bbox variation compared to VisDrone2019-MOT, as evidenced in Fig. \ref{Challenges}. This characteristic implies that UAVDT is more challenging in terms of both detection and embedding tasks. When evaluated on the official server, our results for the UAVDT benchmarks can be seen in Table \ref{table:Quantitative comparisons}. 
DroneMOT continues to set the benchmark, achieving an unrivaled IDF1 score of 69.6\% and a commendable MOTA of 50.1\%. 
Additionally, DroneMOT outperforms by registering a minimal 129 ID switches, underscoring its expertise in consistently preserving object identities across sequences. 

\begin{table}[]
\centering
\caption{Abalation study on Visdrone2019-MOT validation set.}
\label{tab:abalation}
\resizebox{0.98\linewidth}{!}{
\begin{tabular}{cccccc}
\hline
Baseline   & DIA        & MDA        & MOTA(\%) & IDs  & IDF1(\%) \\ \hline
\checkmark &            &            & 29.7     & 1509  & 38.3     \\
\checkmark & \checkmark &            & 33.4     & 1407  & 45.1     \\
\checkmark &            & \checkmark & 32.4     & 406  & 48.9     \\
\checkmark & \checkmark & \checkmark & 34.3     & 218  & 53.4     \\ \hline
\end{tabular}%
}
\end{table}

\begin{table}[]
\centering
\caption{Analysis of the effectiveness of MDA module. The baseline uses the Kalman filter and EMA to update the feature.}
\label{tab:abalation_MDA}
\resizebox{0.98\linewidth}{!}{%
\begin{tabular}{cccccc}
\hline 
Motion model & Appearance model & IDs & IDF1 & IDP & IDR \\ \hline
-            & -                & 1407   & 45.1    & 48.6   & 42.1   \\
DMP          & -                & 229   & 52.8    & 57.8   & 48.6   \\
-            & AFS              & 690   & 46.5    & 52.8   & 41.5   \\
DMP          & AFS              & 218   & 53.4    & 43.0     & 52.8   \\ \hline
\end{tabular}%
}
\vspace{-0.6cm}
\end{table}

\subsection{Ablation Study}
The baseline model we compared against is FairMOT\cite{fairmot}, which uses DLA34 as its backbone and has the same loss settings as DroneMOT.

\begin{figure}[t] 
        \centering
        \setlength{\abovecaptionskip}{-0.6cm}
        \includegraphics[width=3.4in]{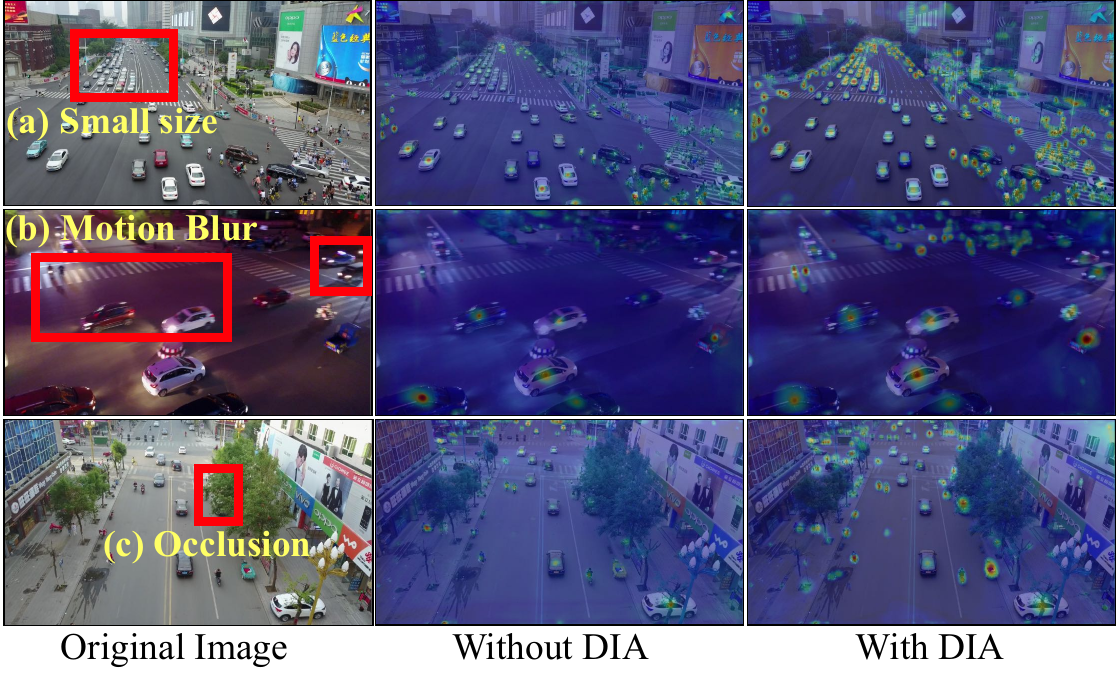}
        \caption{Feature map comparison between without DIA and with DIA.} 
        \label{fig:DIA_visual}
        \vspace{-0.3cm}
\end{figure}

\noindent
\textbf{Dual-Domain Integrated Attention.}
The DIA module, powered by spatial attention and heatmap-guided temporal attention, significantly refines feature representation, bolstering robustness and accuracy. As evidenced in Table~\ref{tab:abalation}, including the DIA module enhances the MOTA and IDF1 scores to $20.4\%$ and $45.1\%$, respectively. 
Furthermore, it results in a decrease in IDs, dropping from 1509 to 1407. The proficiency of the DIA module is visually represented in Fig.~\ref{fig:DIA_visual}, which underscores its effectiveness in assisting the network to recognize small-sized, blurred, or occluded objects.

\noindent
\textbf{Motion-Driven Association Module.} 
The integration of the MDA module plays a pivotal role in enhancing tracking performance, as evident in Table \ref{tab:abalation}. Specifically, we observe improvements of $4.7\%$ in MOTA and $10.6\%$ in IDF1. Moreover, IDs are significantly reduced, plummeting from 1509 to 406. Delving deeper into the MDA module's components in Table \ref{tab:abalation_MDA}, we find that the DMP component substantially curtails ID switches, bringing them down from 1407 to 229. Further synergizing DMP with AFS elevates the IDF1 score to $53.4\%$, underscoring the combined strength of both components in refining tracking accuracy.

\begin{figure}[t] 
        \centering
        \setlength{\abovecaptionskip}{-0.6cm}
        \includegraphics[width=3.4in]{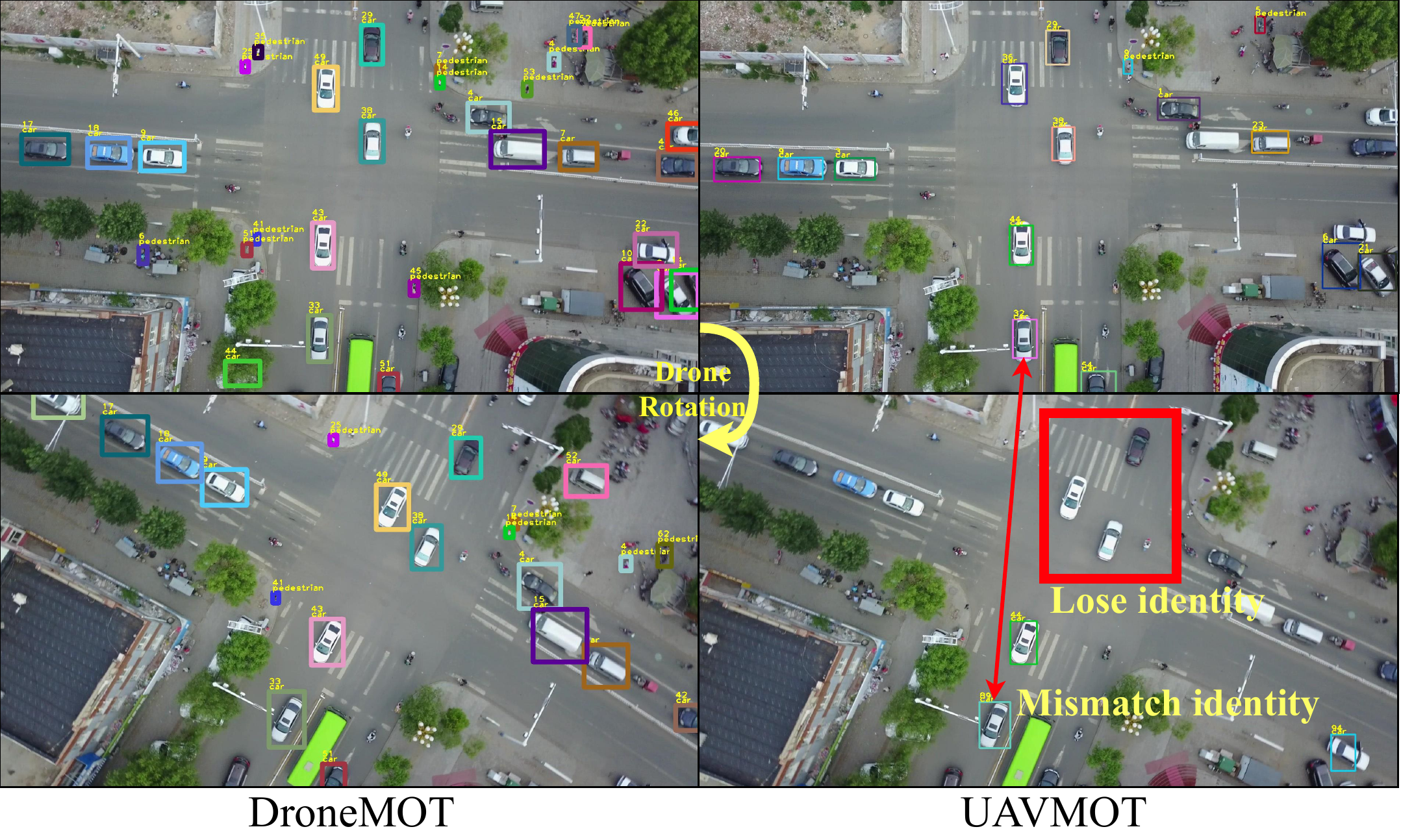}
        \caption{Visualization of tracking results on the \textbf{Visdrone2019-MOT} dataset when the drone is rotating rapidly.} 
        \label{fig:visual_visdrone}
        \vspace{-0.3cm}
\end{figure}\begin{figure}[!h] 
        \centering
        \setlength{\abovecaptionskip}{-0.6cm}
        \includegraphics[width=3.4in]{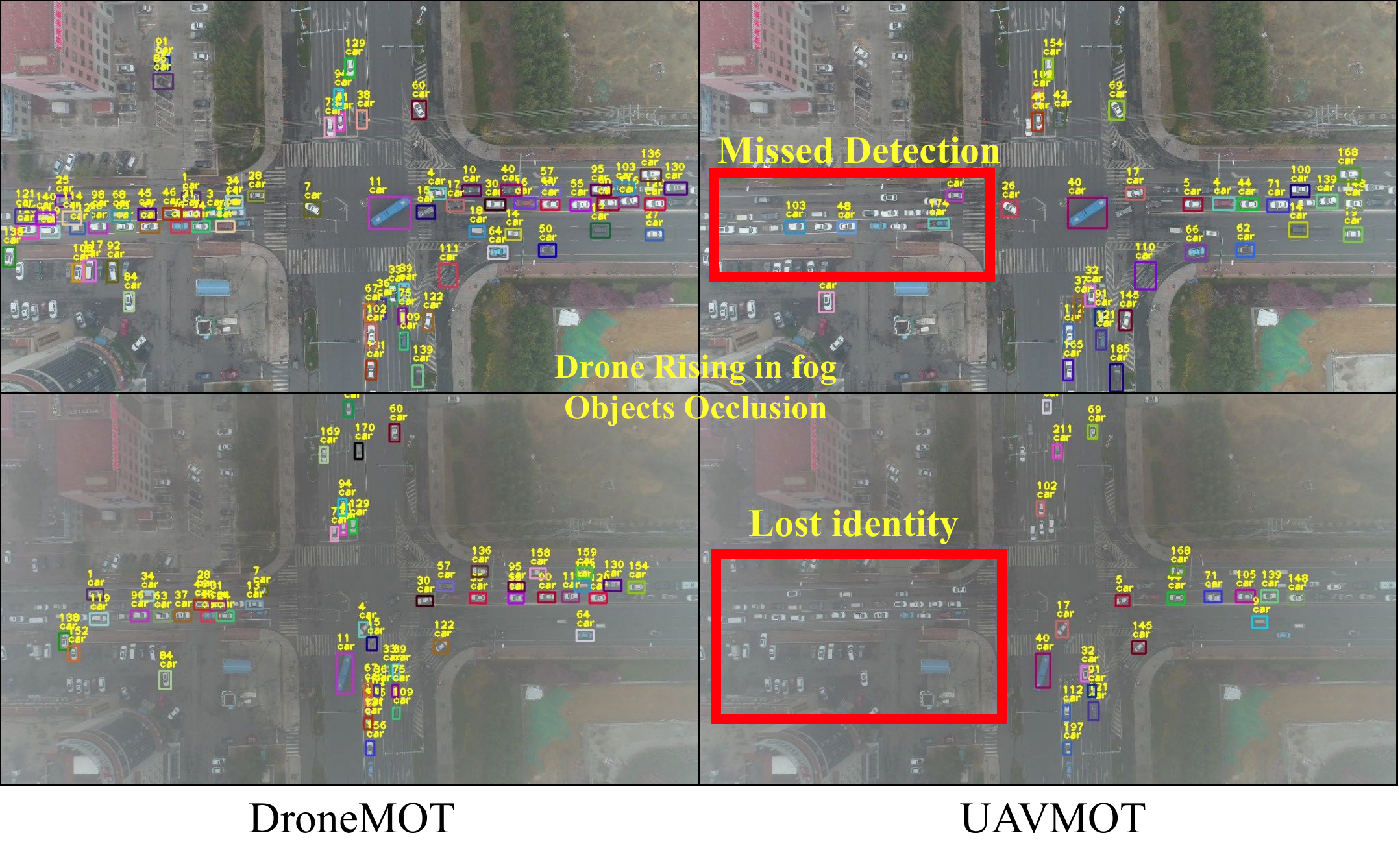}
        \caption{Visualization of tracking results on the \textbf{UAVDT} dataset when the drone raises in foggy conditions, and the target is obscured by the fog.} 
        \label{fig:visual_uavdt}
        \vspace{-0.6cm}
\end{figure}

\subsection{Visualization}
To showcase the efficacy of DroneMOT, we present a tracking visualization compared to UAVDT. 
Particularly during drone rotations, DroneMOT consistently retains the trajectory ID of targets, ensuring no loss or mismatch of IDs, as evidenced in Fig.~\ref{fig:visual_visdrone}. 
Even under challenging foggy conditions, exemplified in Fig.~\ref{fig:visual_uavdt}, DroneMOT's DIA module proves instrumental in accurately identifying targets — even the minute ones obscured by fog cover as the drone ascends. These visual representations highlight how adeptly DroneMOT adapts to diverse and dynamic conditions, excelling in the MOT task on drone footage.

\section{CONCLUSIONS}
In this paper, we introduced DroneMOT, a novel approach tailored specifically for the challenges presented by drone-based multiple object tracking. 
By integrating the proposed Dual-Domain Integrated Attention, DroneMOT excels in object detection and feature embedding, capitalizing on spatial nuances and leveraging heatmap-guided temporal insights. 
Moreover, our Motion-Driven Association scheme delivers a robust data association method, recognizing the combined movement of drones and objects. This is further enriched by our innovative Adaptive Feature Synchronization (AFS) and Dual Motion-based Prediction modules. 
Empirical results validate DroneMOT's superiority over existing methods for drone-based MOT.


\bibliographystyle{IEEEtran}
\bibliography{References}

\end{document}